\documentclass[10pt,twocolumn,letterpaper]{article}

\usepackage[pagenumbers]{cvpr} % To force page numbers, e.g. for an arXiv version

\usepackage{algorithm}
\usepackage{algorithmic}
\definecolor{cvprblue}{rgb}{0.21,0.49,0.74}
\usepackage[pagebackref,breaklinks,colorlinks,allcolors=cvprblue]{hyperref}

\def\paperID{*****} % *** Enter the Paper ID here
\def\confName{CVPR}
\def\confYear{2026}

\title{Selective Mask Propagation for Multi-Object Tracking}

\author{Alexander Holmberg\\
KTH Royal Institute of Technology \\
{\tt\small alholmbe@kth.se}
}

\begin{document}
\maketitle

% ============================================================
% ABSTRACT
% ============================================================
\begin{abstract}
In multi-object tracking, most frames are easy for a lightweight base tracker while a small fraction is intrinsically hard. Video object segmentation (VOS) models can often preserve identity through the hard frames where the base tracker fails, but they are much more expensive in compute and memory. We propose selective mask propagation, a tracking algorithm that dispatches from a base tracker to a VOS model only on windows where an assignment-uncertainty signal fires. The base tracker's output is modified only when the VOS model makes a confident prediction that contradicts the base tracker's identity assignment; weak or inconclusive predictions preserve the base output. The method is training-free, treats both the base tracker and the VOS model as black boxes, and can benefit from replacing the VOS component with a more capable model. On DanceTrack, selective mask propagation significantly improves three different base trackers. On SportsMOT, where identity preservation is central to sports analytics, SAM~3-Deep-EIoU with global track association achieves state-of-the-art performance on the benchmark with 87.2 HOTA.
\end{abstract}

% ============================================================
% 1. INTRODUCTION
% ============================================================
\section{Introduction}

Multi-object tracking (MOT) requires maintaining consistent identities across the frames of a video. Modern tracking-by-detection methods~\cite{zhang2022bytetrackmultiobjecttrackingassociating, cao2023observationcentricsortrethinkingsort, huang2023iterativescaleupexpansionioudeep} build a cost matrix from spatial, motion, or appearance cues and solve the resulting assignment problem with the Hungarian algorithm. This works when one assignment is clearly better than the alternatives. It fails when the alternatives are near-tied: when two people are close in the cues the tracker uses, candidate assignments have nearly identical cost and the tracker commits to one identity on a small cost difference. When the wrong choice is made, the result is an identity switch that propagates forward in time. In sports analytics, this can corrupt every subsequent event attributed to that trajectory, assigning passes, shots, or defensive actions to the wrong player until the identity is recovered.

Existing approaches address this failure mode at different costs. End-to-end trackers~\cite{zeng2022motrendtoendmultipleobjecttracking, zhang2023motrv2bootstrappingendtoendmultiobject, gao2024memotrlongtermmemoryaugmentedtransformer} learn detection and association jointly, removing the explicit cost matrix, but require extensive training data and do not transfer cleanly across domains without retraining. SORT-family extensions~\cite{du2023strongsortmakedeepsortgreat, aharon2022botsortrobustassociationsmultipedestrian, yang2024hybridsortweakcuesmatter} improve the cost matrix with better motion or appearance features, but cannot resolve cases where those cues are simultaneously ambiguous. Mask-based trackers~\cite{stanczyk2024temporallypropagatedmasksbounding, cheng2023trackingdecoupledvideosegmentation, li2024matchingsegmenting} integrate video object segmentation as an additional association cue, but propagate masks on every frame, paying the model's cost regardless of whether the base tracker needs help.

Video object segmentation (VOS) models such as the SAM family~\cite{ravi2024sam2segmentimages, carion2025sam3segmentconcepts} and Cutie~\cite{cheng2024puttingobjectvideoobject} target this failure mode directly. They maintain a pixel-level representation of each tracked object and can preserve identity through occlusions where bounding-box association fails. Bringing them into MOT raises two practical obstacles. First, per-frame latency and GPU memory grow with the number of tracked objects. Second, uniform VOS usage subjects every frame to mask drift, mask collapse, or mask convergence. On easy frames where a lightweight tracker would already have been correct, these failures degrade the output rather than improve it.

\begin{figure}[t]
\centering
\includegraphics[width=\columnwidth]{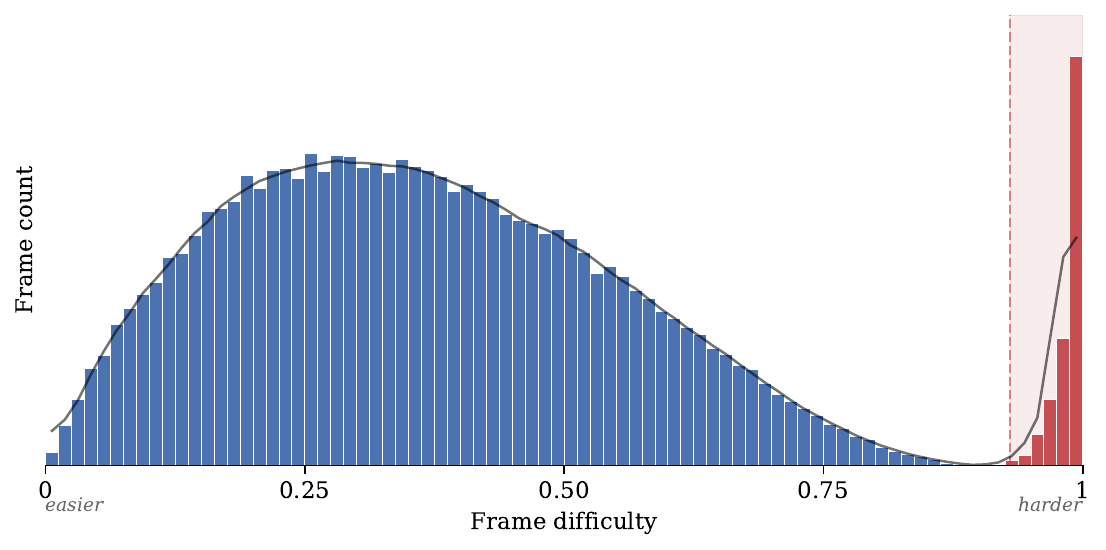}
\caption{Schematic illustration of the distribution of frame difficulty in multi-object tracking. Most frames are easy for lightweight bounding-box association, while a small fraction concentrates the ambiguity that produces identity switches. Section~\ref{sec:window_creation} introduces the assignment margin, the measurable quantity used to identify the hard fraction at runtime.}
\label{fig:frame_difficulty}
\end{figure}

The structure of the problem suggests a division of labor. Most frames are easy for a lightweight tracker, while a small fraction is intrinsically hard (Figure~\ref{fig:frame_difficulty}). The cost of running a more capable model is roughly constant per frame, but its value is concentrated almost entirely in the hard fraction. We therefore propose selective mask propagation: the base tracker handles the common case, and a VOS model is dispatched only to temporal windows where a signal indicates that the base tracker's identity assignment is at risk.

The primary signal is the assignment margin in the Hungarian cost matrix, defined as the difference between the second-best and best assignment cost for a matched detection. A large margin indicates that one assignment is clearly preferred; a small margin indicates that competing assignments are near-tied. When the signal fires, the VOS model propagates masks through the ambiguous window. At exit, the base tracker's output is modified only when the VOS model makes a confident prediction that contradicts the base tracker's identity assignment. If the VOS output is weak, inconclusive, or confirms the base tracker, the original output passes through unchanged. This asymmetric decision rule restricts the algorithm's relative failure modes: the only way it can harm the base tracker is through a false-positive identity switch, while all other outcomes preserve the base output.

We instantiate selective mask propagation as SAM-Deep-EIoU, using Deep-EIoU~\cite{huang2023iterativescaleupexpansionioudeep} as the base tracker and SAM~3~\cite{carion2025sam3segmentconcepts} for mask propagation. For sports applications, we additionally incorporate a global track association (GTA)~\cite{sun2024gtaglobaltrackletassociation} module that links tracklets across frame-boundary exits using jersey recognition~\cite{koshkina2024generalframeworkjerseynumber}, team classification, and appearance embeddings. Our contributions are:
\begin{itemize}
    \item A training-free tracking algorithm that selectively dispatches from a lightweight base tracker to a VOS model using assignment uncertainty, while preserving the base tracker's output unless the VOS model confidently contradicts the base assignment.
    \item State-of-the-art performance on SportsMOT~\cite{cui2023sportsmotlargemultiobjecttracking} (87.2 HOTA), and consistent improvements across three base trackers on DanceTrack~\cite{sun2022dancetrackmultiobjecttrackinguniform}.
\end{itemize}

% ============================================================
% 2. RELATED WORK
% ============================================================
\section{Related Work}

\noindent\textbf{Tracking-by-detection.}
The dominant paradigm for multi-object tracking builds a cost matrix from spatial, motion, or appearance cues, then solves the assignment with the Hungarian algorithm. ByteTrack~\cite{zhang2022bytetrackmultiobjecttrackingassociating} associates detections by IoU in two rounds. OC-SORT~\cite{cao2023observationcentricsortrethinkingsort} improves motion estimation during occlusion. StrongSORT~\cite{du2023strongsortmakedeepsortgreat} adds re-identification embeddings and camera-motion compensation. Deep-EIoU~\cite{huang2023iterativescaleupexpansionioudeep} combines expanded IoU with appearance embeddings, achieving strong results without Kalman filtering in the association step. These methods differ in how they construct the cost matrix, but all expose the same assignment structure. Selective mask propagation uses that structure by reading assignment ambiguity from the cost matrix rather than modifying the base tracker.

\noindent\textbf{Mask-based tracking.}
Several methods incorporate segmentation masks into tracking. DEVA~\cite{cheng2023trackingdecoupledvideosegmentation} decouples image segmentation from class-agnostic temporal propagation. MASA~\cite{li2024matchingsegmenting} uses SAM segmentations to learn instance-level correspondences for association. McByte~\cite{stanczyk2024temporallypropagatedmasksbounding} is the closest comparison: it integrates temporally propagated masks into ByteTrack as an additional association cue. These methods demonstrate the value of pixel-level representations for identity preservation, but they use masks as a standing part of the tracker. In contrast, selective mask propagation invokes the VOS model only when a dispatch signal fires and accepts its output only as a gated correction to the base tracker.

\noindent\textbf{Video object segmentation.}
VOS models propagate pixel-level masks across video frames. Cutie~\cite{cheng2024puttingobjectvideoobject} uses object-level memory encoding for robust long-term propagation. SAM~2~\cite{ravi2024sam2segmentimages} extends the Segment Anything model to video with a streaming memory architecture supporting prompted segmentation. SAM~3~\cite{carion2025sam3segmentconcepts} further improves mask quality and temporal consistency. These models are powerful for single-object tracking but lack the tracklet management required for MOT, including handling new objects, track termination, and identity assignment.

\noindent\textbf{Global track association.}
Global association operates on finished tracklets rather than on a per-frame basis. In sports, GTA~\cite{sun2024gtaglobaltrackletassociation} links tracklets using stable identity cues such as jersey numbers, team identity, and appearance. This addresses off-screen re-identification after players leave and re-enter the camera view. Selective mask propagation targets the complementary failure mode: on-screen identity switches during occlusion and close interaction.

\noindent\textbf{Selective computation.}
The broader idea of allocating expensive computation selectively appears in early-exit networks such as BranchyNet~\cite{teerapittayanon2017branchynetfastinferenceearly} and MSDNet~\cite{huang2018multiscaledensenetworksresource}, which route easy inputs through cheaper exits and reserve deeper computation for uncertain inputs. Selective mask propagation applies the same principle to tracking-by-detection: a lightweight tracker handles the common case, while a more capable VOS model is dispatched to ambiguous windows.

% ============================================================
% 3. METHOD
% ============================================================
\section{Method}

Selective mask propagation augments a tracking-by-detection base tracker with a VOS model that is invoked only on at-risk temporal windows. The base tracker runs on every frame and supplies the default output. A dispatch signal opens a window when the base tracker's association is ambiguous, the VOS model propagates masks through that window, and the output is modified only if the propagated mask gives a confident prediction that contradicts the base tracker's identity assignment. If the VOS result is weak, inconclusive, or confirms the base tracker, the base output is preserved.

For sports tracking, we additionally separate two identity problems: \emph{on-screen identity preservation}, maintaining identities through occlusions while players remain visible, and \emph{off-screen re-identification}, linking tracklets of players who leave and re-enter the scene. Selective mask propagation addresses the first problem. Global track association (GTA), described in Section~\ref{sec:gta}, addresses the second. Pseudocode for the complete pipeline is given in Appendix~\ref{app:pseudocode}.

\subsection{Window Creation}
\label{sec:window_creation}

The core dispatch question is when to invoke mask propagation. We define signals that identify frames where the base tracker is likely to make an identity error. When a signal fires, a \emph{window} is opened: SAM is seeded from a clean frame before the ambiguity and propagates forward through it.

The primary signal monitors the assignment margin in the Hungarian cost matrix. Given a cost matrix $\mathbf{C} \in \mathbb{R}^{N \times M}$ where $C_{i,j}$ is the cost of assigning track $i$ to detection $j$, let $(i^*, j^*)$ be a matched pair from the Hungarian solution. The margin for detection $j^*$ is:
\begin{equation}
    m_{j^*} = \min_{i \neq i^*} C_{i,j^*} - C_{i^*,j^*}
\end{equation}
A large margin indicates confident assignment; a small margin means two tracks are nearly equally viable for the same detection. The margin is computed from the matrix the Hungarian algorithm already operates on, so it is free at runtime and requires no addition to the base tracker. We open a window when $m_{j^*} < \tau_{\text{entry}}$ for any matched pair.

Additional signals extend coverage to cases where the margin signal is absent or incomplete. A gap signal fires when a track disappears for at least $G$ frames and reappears, since the track may have been re-attached to the wrong person after an occlusion. Witness windows are opened for nearby tracks that overlap a primary window at entry, since the nearby track is often the counterpart in an identity switch and both identities must be tracked to resolve the ambiguity.

\noindent\textbf{Seeding.}
Each window requires a seed frame before the ambiguity where the target identity is unambiguous and spatially isolated. We walk backward from the entry frame to find $N_{\text{seed}}$ consecutive frames with margin $\geq \tau_{\text{seed}}$ and no nearby overlapping boxes. The high margin confirms assignment confidence; the isolation ensures SAM receives unambiguous pixels for initialization. Windows without a valid seed are discarded.

The entry threshold $\tau_{\text{entry}}$ controls a compute-versus-recall tradeoff. A more relaxed threshold opens more windows. Most additional windows produce non-SWAP outcomes that preserve the base tracker, but some catch real identity switches that a stricter threshold would miss. False-positive SWAPs remain the possible relative failure mode, so the threshold controls both the amount of VOS compute and the number of opportunities for the VOS model to intervene.

\begin{figure*}[t]
\centering
\includegraphics[width=\textwidth]{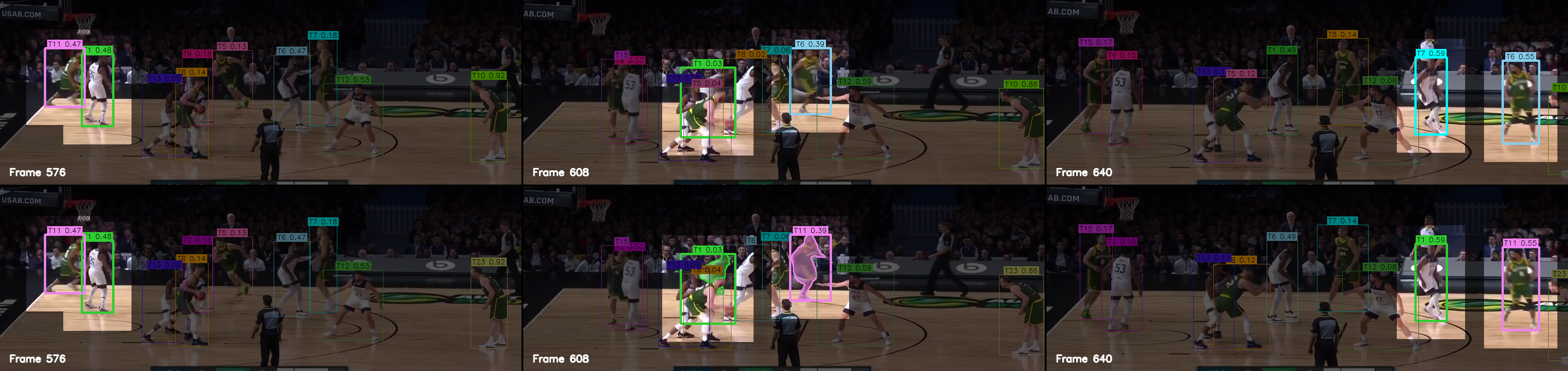}
\caption{Qualitative example from SportsMOT. Two players (T1, T11) are involved in an occlusion event during a play. \textbf{Top:} Deep-EIoU loses both identities after the occlusion, assigning T1 and T11 to the wrong people at exit (T1$\to$T7, T11$\to$T6). \textbf{Bottom:} SAM-Deep-EIoU seeds masks before the occlusion (left), propagates through it (center, masks shown), and correctly resolves both identities at exit (right).}
\label{fig:qualitative}
\end{figure*}

\subsection{Mask Propagation and Exit Conditions}
\label{sec:propagation}

\begin{figure}[t]
\centering
\includegraphics[width=\columnwidth]{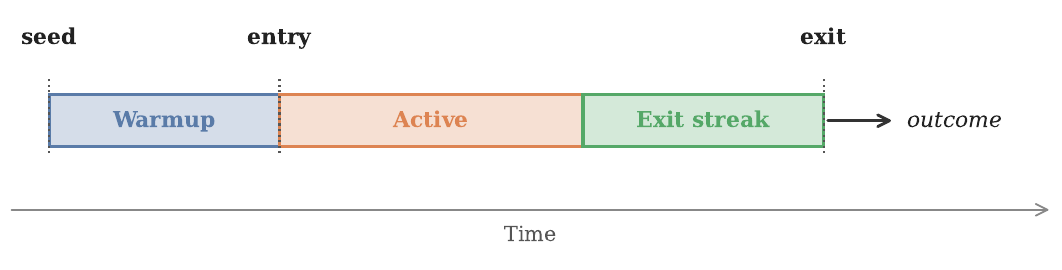}
\caption{A window's lifetime. The VOS model propagates from the seed frame, exit checks begin at the entry frame, and the window exits once all four exit conditions have held for $N_{\text{exit}}$ consecutive frames.}
\label{fig:window_lifetime}
\end{figure}

Once windows are created and seeded, masks are propagated forward from the seed frame through the ambiguous region. Figure~\ref{fig:window_lifetime} summarizes a window's lifetime from seed to exit.

To determine when the ambiguity has resolved, we check which base tracker bounding box the mask has settled into using intersection over mask area (IoMA):
\begin{equation}
    \text{IoMA}(M, B) = \frac{|M \cap B|}{|M|}
\end{equation}
where $M$ is the set of mask pixels and $B$ is the set of pixels within the bounding box. A window exits when the following conditions are sustained for $N_{\text{exit}}$ consecutive frames: (1) the mask is contained by a single track (IoMA $\geq \tau_{\text{IoMA}}$), (2) that track's assignment margin has recovered above $\tau_{\text{exit}}$, (3) the track's bounding box is spatially isolated from other bounding boxes, and (4) the mask is isolated from other active masks.

When the exit conditions are met, the outcome depends on where the mask ended up relative to where it started. If the mask settled into a different track than it was seeded on, the window exits as SWAP and track IDs are renamed from the swap frame, the frame at which the mask first settled into the matched track. If the mask returned to the same track, the window exits as CLEAN and no modification is made.

Several conditions can terminate a window early without an identity decision. If, at the entry frame, the mask no longer covers the seeded track's bounding box (IoMA $< \tau_{\text{IoMA}}$), the base tracker reassigned the seeded identity between seed and entry, and the window exits as STALE. If two masks converge onto the same person, neither identity is recoverable and both are discarded. If the mask area has significantly collapsed relative to its seed, the mask has degraded and is discarded. If the mask reaches the frame border and disappears, the person left the scene. If propagation reaches the last frame of the sequence with no other condition having fired, the window exits as END. These STALE, DEGRADED, EDGE, and END outcomes leave the base tracker's output untouched.

\subsection{Global Track Association}
\label{sec:gta}

For off-screen re-identification in sports, we link tracklets separated by frame-boundary exits, adapting the tracklet connector of Sun et al.~\cite{sun2024gtaglobaltrackletassociation} with two changes: the appearance tier is preceded by discrete jersey-number and team-identity tiers, and their tracklet splitter is omitted, since selective mask propagation maintains tracklet purity before global association runs. We extract jersey numbers via OCR on pose-guided torso crops~\cite{koshkina2024generalframeworkjerseynumber} and use them alongside team classification and appearance embeddings to match tracklets hierarchically: tracklets with the same jersey number and team are merged first, then remaining tracklets are merged by appearance similarity with vetoes for temporal overlap, opposite-edge transitions, and conflicting identities. Gaps left by merged tracklets are filled with linear bounding box interpolation. GTA operates on finished tracklets and is a sports-specific post-processing module, not part of the general selective mask propagation algorithm.

% ============================================================
% 4. RESULTS
% ============================================================
\section{Results}

We evaluate selective mask propagation in two settings. DanceTrack tests whether the same dispatch mechanism improves different base trackers without sport-specific modules. SportsMOT tests the sports-tracking instantiation, where selective mask propagation is combined with GTA.

The pipeline is deterministic given fixed detections, embeddings, and VOS model weights, so the uncertainty in a comparison comes from the finite sample of sequences rather than from run-to-run variation. Validation-split comparisons are therefore paired, with both configurations run on the same sequences with the same detections, and we report two-sided Wilcoxon signed-rank $p$-values computed over the nonzero per-sequence HOTA deltas. Test-set comparisons against published methods are point comparisons under the benchmark's single-submission protocol.

\subsection{Cross-Tracker Generalization on DanceTrack}

To validate that selective mask propagation generalizes across base trackers, we evaluate on the DanceTrack~\cite{sun2022dancetrackmultiobjecttrackinguniform} validation set using three base trackers with the same YOLOX detections. No GTA or sport-specific module is used; the only addition to each base tracker is the margin-dispatched mask propagation described in Section~\ref{sec:window_creation}--\ref{sec:propagation}.

\begin{table}[h]
\centering
\small
\begin{tabular}{l|ccc}
\hline
Base tracker & Baseline & + SAM~2 & + SAM~3 \\
\hline
\multicolumn{4}{l}{\textit{HOTA ($\uparrow$)}} \\[-1pt]
\hline \\[-8pt]
SORT & 39.8 & 45.1\,\textsuperscript{\tiny+5.3} & \textbf{46.1}\,\textsuperscript{\tiny+6.2} \\
ByteTrack & 54.6 & 60.3\,\textsuperscript{\tiny+5.7} & \textbf{61.2}\,\textsuperscript{\tiny+6.6} \\
Deep-EIoU & 51.7 & 57.8\,\textsuperscript{\tiny+6.1} & \textbf{59.7}\,\textsuperscript{\tiny+8.0} \\
\hline
\multicolumn{4}{l}{\textit{AssA ($\uparrow$)}} \\[-1pt]
\hline \\[-8pt]
SORT & 23.8 & 30.0\,\textsuperscript{\tiny+6.2} & \textbf{31.1}\,\textsuperscript{\tiny+7.3} \\
ByteTrack & 39.0 & 47.0\,\textsuperscript{\tiny+8.0} & \textbf{48.3}\,\textsuperscript{\tiny+9.3} \\
Deep-EIoU & 36.7 & 44.9\,\textsuperscript{\tiny+8.2} & \textbf{47.7}\,\textsuperscript{\tiny+10.9} \\
\hline
\end{tabular}
\caption{DanceTrack validation set (25 sequences). Selective mask propagation applied to three base trackers using the same detections with no per-tracker tuning.}
\label{tab:general}
\end{table}

Table~\ref{tab:general} shows consistent improvements across all base trackers, indicating that the margin signal is not specific to one cost-matrix construction. Every baseline comparison is significant (worst case $p = 5.3 \times 10^{-5}$), and no base tracker degrades on more than 2 of the 25 sequences. The improvement is largest on AssA, directly reflecting the method's purpose: correcting association errors. Upgrading the VOS model from SAM~2 to SAM~3 improves the point estimate for every base tracker with the surrounding algorithm held fixed, significantly for SORT ($p = 0.010$) and Deep-EIoU ($p = 0.003$), while the ByteTrack difference does not reach significance ($p = 0.076$). The algorithm therefore benefits from VOS quality improvements without retraining, supporting the modular design.

\begin{table}[h]
\centering
\small
\begin{tabular}{c|cccc}
\hline
$\tau_{\text{entry}}$ & $\Delta$HOTA & Windows & SWAPs & SAM s/frame \\
\hline
0.010 & +5.9 & 1006 & 117 & 0.105 \\
0.025 & +8.1 & 1200 & 136 & 0.107 \\
0.050 & +8.0 & 1374 & 141 & 0.114 \\
\hline
\end{tabular}
\caption{Effect of $\tau_{\text{entry}}$ on DanceTrack validation set with Deep-EIoU + SAM~3. Relaxing the threshold opens more windows and catches more SWAP outcomes. Timing on a single RTX 5090.}
\label{tab:margin}
\end{table}

Table~\ref{tab:margin} shows the effect of varying $\tau_{\text{entry}}$. The strictest threshold (0.010) reduces compute but misses 24 identity switches relative to the most relaxed setting, and its accuracy is significantly lower than the middle setting ($+2.2$ HOTA from 0.010 to 0.025, $p = 0.013$). Accuracy saturates by 0.025: relaxing further to 0.050 leaves accuracy statistically unchanged ($-0.1$ HOTA, $p = 0.72$) while catching 5 additional switches at marginal extra compute, so the remaining benefit of the relaxed setting is swap recall. We use $\tau_{\text{entry}} = 0.05$ throughout the remaining experiments.

\subsection{SportsMOT}

We evaluate the sports-tracking instantiation (Deep-EIoU + selective mask propagation + GTA) on the SportsMOT~\cite{cui2023sportsmotlargemultiobjecttracking} test set. Table~\ref{tab:sportsmot_sota} compares against published methods.

\begin{table}[h]
\centering
\small
\begin{tabular}{l|cccc}
\hline
Method & HOTA & AssA & IDF1 & MOTA \\
\hline
CenterTrack~\cite{zhou2020trackingobjectspoints} & 62.7 & 48.0 & 60.0 & 90.8 \\
ByteTrack~\cite{zhang2022bytetrackmultiobjecttrackingassociating} & 64.1 & 52.3 & 71.4 & 95.9 \\
MeMOTR~\cite{gao2024memotrlongtermmemoryaugmentedtransformer} & 70.0 & 59.1 & 71.4 & 91.5 \\
OC-SORT~\cite{cao2023observationcentricsortrethinkingsort} & 73.7 & 61.5 & 74.0 & 96.5 \\
MotionTrack~\cite{qin2023motiontracklearningrobustshortterm} & 74.0 & 61.7 & 74.0 & 96.6 \\
DiffMOT~\cite{lv2024diffmotrealtimediffusionbasedmultiple} & 76.2 & 65.1 & 76.1 & 97.1 \\
Deep-EIoU~\cite{huang2023iterativescaleupexpansionioudeep} & 77.2 & 67.7 & 79.8 & 96.3 \\
GTA~\cite{sun2024gtaglobaltrackletassociation} & 81.0 & 74.5 & 86.5 & 96.3 \\
NOOUGAT~\cite{missaoui2025noougatunifiedonlineoffline} & 85.6 & 83.0 & 92.3 & 95.9 \\
\hline
SAM~2-Deep-EIoU & 85.5 & 81.7 & 91.2 & 97.3 \\
\textbf{SAM~3-Deep-EIoU} & \textbf{87.2} & \textbf{84.2} & \textbf{93.6} & \textbf{98.1} \\
\hline
\end{tabular}
\caption{SportsMOT test set. SAM~3-Deep-EIoU achieves state-of-the-art performance across all four metrics shown.}
\label{tab:sportsmot_sota}
\end{table}

SAM~3-Deep-EIoU achieves 87.2 HOTA on SportsMOT, state-of-the-art across all four metrics shown. GTA~\cite{sun2024gtaglobaltrackletassociation} applies global tracklet association on top of the same Deep-EIoU base tracker and reaches 81.0 HOTA, so the margin over their result is attributable to selective mask propagation and our variant of the association module rather than to a stronger tracking-by-detection core. SAM~3 outperforms SAM~2 by +1.7 HOTA with the dispatch signal, window logic, and merge procedure held fixed, a point comparison consistent with the paired DanceTrack evidence above.

\begin{table}[h]
\centering
\small
\begin{tabular}{l|ccc}
\hline
Configuration & HOTA & AssA & IDF1 \\
\hline \\[-8pt]
Deep-EIoU (DE) & 78.9 & 70.7 & 81.3 \\
DE + GTA & 86.7 & 83.6 & 93.3 \\
SDE & 80.4 & 73.2 & 83.0 \\
\textbf{SDE + GTA} & \textbf{88.5} & \textbf{87.0} & \textbf{95.7} \\
\hline
\end{tabular}
\caption{Component ablation on SportsMOT validation set with SAM~3. SDE denotes Deep-EIoU with selective mask propagation.}
\label{tab:gta_ablation}
\end{table}

Table~\ref{tab:gta_ablation} isolates the two components on the validation set. Selective mask propagation adds +1.5 HOTA over Deep-EIoU ($p = 3.3 \times 10^{-5}$) and +1.8 over DE + GTA ($p = 3.2 \times 10^{-5}$), while GTA adds +7.8 and +8.1 in the corresponding comparisons, improving all 45 sequences in both cases. The combined improvement (+9.6) approximately equals the sum of the individual contributions, consistent with the two components addressing disjoint error modes. GTA's larger share reflects the structure of SportsMOT, where field-of-view exits are the dominant association error and every off-screen re-identification is routed to GTA by design.

\begin{table}[h]
\centering
\small
\begin{tabular}{l|ccc}
\hline
Sport & HOTA & AssA & IDF1 \\
\hline \\[-8pt]
Basketball & \textbf{90.4}\,\textsuperscript{\tiny+9.6} & \textbf{89.5}\,\textsuperscript{\tiny+17.5} & \textbf{97.8}\,\textsuperscript{\tiny+14.2} \\
Football & 87.9\,\textsuperscript{\tiny+12.0} & 85.6\,\textsuperscript{\tiny+18.6} & 94.8\,\textsuperscript{\tiny+19.0} \\
Volleyball & 87.2\,\textsuperscript{\tiny+7.2} & 85.8\,\textsuperscript{\tiny+12.8} & 94.5\,\textsuperscript{\tiny+10.1} \\
\hline
\end{tabular}
\caption{Per-sport breakdown on SportsMOT validation set with SAM~3-Deep-EIoU and GTA. Superscript values are deltas over Deep-EIoU, measuring the contribution of the complete system.}
\label{tab:persport}
\end{table}

Table~\ref{tab:persport} breaks the full system's performance down by sport. The system improves every sequence of every sport over the Deep-EIoU baseline ($p = 6.1 \times 10^{-5}$ per sport, the smallest value attainable with 15 sequences), with HOTA gains between +7.2 and +12.0. Differences between sports are not statistically separable at this sample size.

\begin{figure}[h]
\centering
\includegraphics[width=\columnwidth]{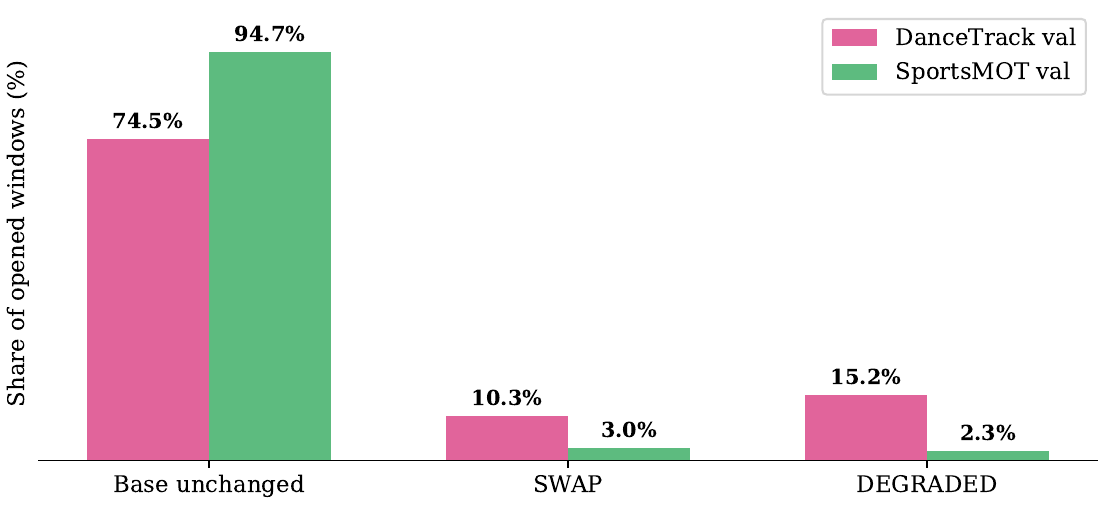}
\caption{Window outcome distribution with SAM~3 and $\tau_{\text{entry}} = 0.05$. Shares are computed over all opened windows (1374 on DanceTrack, 1787 on SportsMOT). Base unchanged groups CLEAN, STALE, EDGE, and END; SWAP is the only outcome that relabels the base tracker's identities; and DEGRADED windows are discarded without modifying the base output.}
\label{fig:outcomes}
\end{figure}

Figure~\ref{fig:outcomes} shows the window outcome distribution across both benchmarks. On both datasets, most windows leave the base tracker's output unchanged. DanceTrack produces a higher rate of both SWAP outcomes and mask degradation, consistent with its more challenging appearance and motion characteristics. The higher degradation rate represents additional compute but does not affect tracking quality, since DEGRADED windows leave the base tracker's output unchanged.

% ============================================================
% 5. CONCLUSION
% ============================================================
\section{Conclusion}

We presented selective mask propagation, a tracking algorithm that combines the efficiency of a lightweight base tracker with the identity-preservation capabilities of a VOS model. The assignment margin in the Hungarian cost matrix provides a dispatch signal for opening at-risk windows, and the asymmetric decision rule preserves the base tracker's output unless the VOS model makes a confident prediction that contradicts the base assignment. The method is training-free, treats both the base tracker and the VOS model as black boxes, and significantly improves SORT, ByteTrack, and Deep-EIoU on DanceTrack without per-tracker tuning.

For sports tracking, SAM~3-Deep-EIoU combines selective mask propagation with global track association, separating on-screen identity preservation from off-screen re-identification. The resulting system achieves state-of-the-art performance on SportsMOT with 87.2 HOTA. The primary limitation is the additional compute cost of running a VOS model on opened windows. Since most opened windows leave the base output unchanged (94.7\% on SportsMOT and 74.5\% on DanceTrack), a more precise dispatch signal could reduce VOS compute while preserving the same identity-switch recall.

{
    \small
    \bibliographystyle{ieeenat_fullname}
    \bibliography{main}
}

% ============================================================
% APPENDIX
% ============================================================
\clearpage
\appendix

\section{Algorithm Pseudocode}
\label{app:pseudocode}

Algorithm~\ref{alg:smp} summarizes the complete selective mask propagation pipeline, with \textsc{PropagationPass} expanded in Algorithm~\ref{alg:propagation}. The named subroutines map to the corresponding subsections of Section~3: \textsc{MarginSignal}, \textsc{GapSignal}, \textsc{WitnessWindows}, and \textsc{SeedSearch} are described in Section~\ref{sec:window_creation}, and the matching, exit checks, and outcomes in \textsc{PropagationPass} are described in Section~\ref{sec:propagation}.

\begin{algorithm}[h]
\caption{Selective Mask Propagation}
\label{alg:smp}
\begin{algorithmic}[1]
\REQUIRE Video $V$, base tracker $B$, VOS model $F$
\ENSURE Tracklets with consistent identities
\STATE $T \gets B(V)$ \COMMENT{base tracker output}
\STATE $W \gets \textsc{MarginSignal}(T) \cup \textsc{GapSignal}(T)$
\STATE $W \gets W \cup \textsc{WitnessWindows}(W, T)$
\FORALL{$w \in W$}
    \STATE $w.\text{seed} \gets \textsc{SeedSearch}(w, T)$
\ENDFOR
\STATE $W \gets \{\, w \in W : w.\text{seed} \neq \emptyset \,\}$ \COMMENT{discard invalid windows}
\STATE $L \gets \textsc{PropagationPass}(W, T, F)$ \COMMENT{rename log}
\STATE $T \gets \textsc{ApplyRenames}(T, L)$
\STATE $T \gets \textsc{GTA}(T)$ \COMMENT{sports configuration}
\STATE \textbf{return} $T$
\end{algorithmic}
\end{algorithm}

\begin{algorithm}[h]
\caption{\textsc{PropagationPass}}
\label{alg:propagation}
\begin{algorithmic}[1]
\REQUIRE Windows $W$, tracklets $T$, VOS model $F$
\ENSURE Rename log $L$
\STATE $L \gets \emptyset$
\STATE $A \gets \emptyset$ \COMMENT{active windows}
\FOR{frame $f$ from $\min_{w \in W} w.\text{seed}$ to end of video}
    \FORALL{$w \in W$ with $w.\text{seed} = f$}
        \STATE Seed $F$ with $w$'s track bounding box at $f$
        \STATE $A \gets A \cup \{w\}$
    \ENDFOR
    \STATE Propagate masks via $F$ to frame $f$
    \STATE Remove converged mask pairs from $A$ \COMMENT{DEGRADED}
    \FORALL{$w \in A$ with $f \geq w.\text{entry}$}
        \STATE Match $F(w)$ to a base tracker box (Sec.~\ref{sec:propagation})
        \STATE Append $(f, \text{matched track})$ to $w$'s match history
        \STATE $o \gets \textsc{ExitCheck}(w, f)$ \COMMENT{Sec.~\ref{sec:propagation}}
        \IF{$o = $ SWAP}
            \STATE Append SWAP record to $L$
        \ENDIF
        \IF{$o$ is terminal}
            \STATE $A \gets A \setminus \{w\}$
        \ENDIF
    \ENDFOR
\ENDFOR
\STATE \textbf{return} $L$
\end{algorithmic}
\end{algorithm}

\end{document}